# Translation Word-Level Auto-Completion:
# What can we achieve out of the box?

**Yasmin Moslem**
ADAPT Centre
School of Computing
Dublin City University
Dublin, Ireland
yasmin.moslem@adaptcentre.ie

**Rejwanul Haque**
School of Computing
National College of Ireland
Mayor Street, IFSC
Dublin, Ireland
rejwanul.haque@ncirl.ie

**Andy Way**
ADAPT Centre
School of Computing
Dublin City University
Dublin, Ireland
andy.way@adaptcentre.ie

## Abstract

Research on Machine Translation (MT) has achieved important breakthroughs in several areas. While there is much more to be done in order to build on this success, we believe that the language industry needs better ways to take full advantage of current achievements. Due to a combination of factors, including time, resources, and skills, businesses tend to apply pragmatism into their AI workflows. Hence, they concentrate more on outcomes, e.g. delivery, shipping, releases, and features, and adopt high-level working production solutions, where possible. Among the features thought to be helpful for translators are sentence-level and word-level translation auto-suggestion and auto-completion. Suggesting alternatives can inspire translators and limit their need to refer to external resources, which hopefully boosts their productivity. This work describes our submissions to WMT's shared task on word-level auto-completion, for the Chinese-to-English, English-to-Chinese, German-to-English, and English-to-German language directions. We investigate the possibility of using pre-trained models and out-of-the-box features from available libraries. We employ random sampling to generate diverse alternatives, which reveals good results. Furthermore, we introduce our open-source API, based on CTranslate2, to serve translations, auto-suggestions, and auto-completions.

## 1 Introduction

Translation auto-suggestion and auto-completion are among the important features that can help translators better utilize Machine Translation (MT) systems. In a Computer-Aided Translation (CAT) environment, a translator can make use of the MT word auto-suggestion feature as follows:

- typing a few words, or clicking a word in a proposed MT translation, a list of suggestions is displayed, as illustrated by Figure 1.

- selecting one of the word suggestions from the list, the rest of the translation is modified accordingly.

The WMT's Word-Level AutoCompletion (WLAC) shared task addresses a more specific scenario, where the user types a few characters, and the system predicts and auto-completes the correct word, given the current context. The WLAC task even suggests that the context might be partial, and it can consist of preceding and/or following words. Given a source sequence $x$, typed character sequence $s$ and a context $c$, WLAC aims to predict a target word $w$ which is to be placed in the middle between the left context $c_l$ and right context $c_r$ to constitute a partial translation. Note that the last word of $c_l$, the auto-completed word $w$, and the first word of $c_r$ are not necessary consecutive.

Previous work proposed diverse approaches, mostly to translation sentence-level auto-suggestion and auto-completion. In their work, Li et al. (2021) proposed an approach to tackle the word-level auto-completion task. Given a tuple $(x, c, s)$, the system decomposes the word autocompletion process into two parts: 1) model the distribution of the target word $w$ based on the source sequence $x$ and the translation context $c$; and 2) find the most possible word $w$ based on the distribution and human typed sequence $s$. Hence, they first use a single placeholder [MASK] to represent the unknown target word $w$, and use the representation of [MASK] learned from the word prediction model, based on BERT (Devlin et al., 2019), to predict it. Then, the predicted distribution of the masked token is used over the vocabulary to filter out invalid words, namely those that do not start with the human typed sequence $s$.

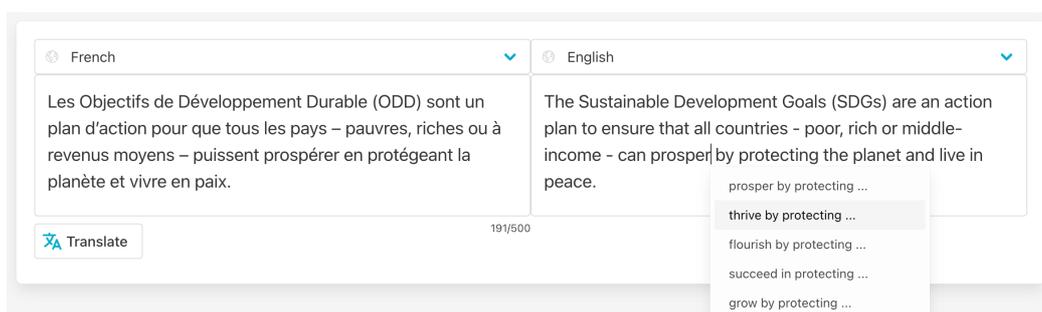

Figure 1: Auto-Suggest: Word Suggestions List[1]

Finally, they return the token with the highest probability over the new distribution.

Researchers in other natural language processing areas such as language modelling offered approaches to improve predictions of decoder-only autoregressive models, trained to predict the next word given the previous context. Among these approaches are top-K sampling and top-p (nucleus) sampling (Fan et al., 2018; Holtzman et al., 2018; Radford et al., 2019; Holtzman et al., 2020). Since neural machine translation inference depends on a decoder model, such approaches from language modelling can be employed. In particular, we investigate the use of top-K sampling during decoding to generate better word-level auto-completions.

## 2 User Survey

Previous work reported that a user can save over 60% of the keystrokes needed to produce a translation in a word completion scenario (Langlais et al., 2000). Other researchers noted that post-editing is faster than MT auto-completion (Koehn, 2009) while MT auto-completion can yield higher quality translation when the baseline MT quality is high (Green et al., 2014).

In a user survey we designed and distributed via social media networks, we asked participants whether they thought an MT word-level auto-suggestions feature could be helpful, and provided a simple definition and an illustrative image. If their answer was "yes", the respondent was asked to specify a reason. By the time of writing this paper, we received 41 responses to our survey. While we do not believe this survey is enough to justify introducing an auto-suggestions feature for every MT system, it can be an indicator as to why some users think such a feature could be helpful. To answer the question, "Which of the following best describes you?" 46.3% (19) of the respondents chose "Translator/Linguist", 31.7% (13) selected "NLP Engineer/Researcher", and the rest 22% (9) were other "MT Users", not included in the two aforementioned categories.

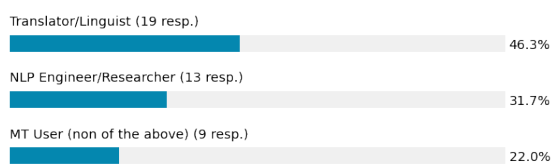

Figure 2: MT user categories

Among the respondents to the survey, 90.2% (37) answered "Yes" to the question "In general, do you believe that a word-level auto-suggestions feature is helpful?". Figure 3 shows the breakdown of answers to the question, "Why do you believe that a word-level auto-suggestions feature can be helpful?" taking into consideration those who answered "No" to the previous question.

Out of the 37 persons who believed a word-level auto-suggestions feature can be helpful, 40.5% (15) of the respondents specified that it can give them some inspiration. This answer is specifically interesting as it is not constrained by time-saving benefits; hence, it focusses more on effectiveness rather than efficiency. The respondent that answered with "Other" mentioned that it allows them to look for alternative senses or phrasings, especially when they suspect the initial translation is bad, and referred to this as "human in the loop".

Respondents were allowed to give extra comments; among the notable comments were:

---
[1]The image is from our demo at: https://www.machinetranslation.io/

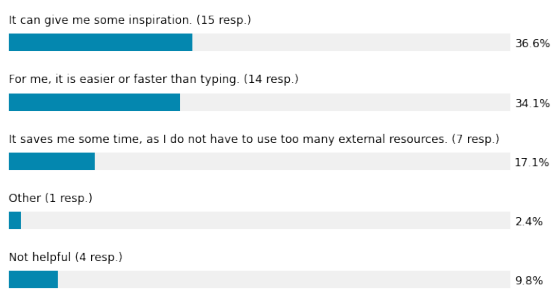

Figure 3: How translators and other MT users perceive word-level auto-suggestions

- *I think word-level suggestions can be a useful feature, particularly when the target language can have several translations of a single source word.*
- *Word-level suggestions can be helpful, but sometimes you end up spending a lot of time figuring out if the MT suggestion is a valid translation in that context. So, I'm not really sure yet how I feel about it.*
- *It's useful, as long as it's seen as a suggestion, and not inserted in the target where the translator is typing.*

Among the respondents who answered with "For me, it is easier or faster than typing", comments included:

- *Though most of the time; the suggestions are lousy.*
- *I don't think it gives me inspiration as I mostly need it for structures, not single words.*
- *Auto-suggestion does not have to come from machine translation. History is much more useful.*

The last comment above might be referring to the fact that in some CAT tools, auto-suggestions can also include glossary terms, and translation memory sub-segments, which encourages further research efforts to investigate methods to enhance leveraging and interaction between various translation resources in human-in-the-loop environments.

We hope this survey will inspire future user studies to look deeper into how diverse users of MT and CAT tools prefer to utilize certain features, such as auto-suggestions, and the value they seek. More aspects should be taken into consideration such as language pairs, translation workflows, and user interfaces. This can help improve these features to better support linguists and other MT users and boost their productivity as well as translation quality.

## 3 Experimental Setup

**Models** We use OPUS pre-trained models[2] based on the Transformer architecture (Vaswani et al., 2017) for the Chinese-to-English, English-to-Chinese, German-to-English, and English-to-German language directions.

**Tokenizers** OPUS models depend on SentencePiece[3] (Kudo and Richardson, 2018) for tokenization. Hence, we use their provided subword models during our pre-processing and post-processing processes. As OPUS's English-to-Chinese model requires defining the target dialect using a pre-specified token, we prepend [">>cmn_Hans<<"] to the list of tokens generated by SentencePiece. For word-level tokenization, we use NLTK for English and German, and Jieba[4] for Chinese. This list of words can be used later to find the word that starts with the typed sequence.

**Inference Engine** We employ CTranslate2 (Klein et al., 2020) for sentence-level MT, as well as for translation auto-suggestions. To this end, we first convert OPUS models into the CTranslate2 format. After that, we utilize a number of CTranslate2 decoding features, including "alternatives at a position" and "auto-completion".[5] The translation options $return\_alternatives$ and $num\_hypotheses$ are essential for all our experiments; the former should be set to $True$ while the latter determines the number of returned alternatives. These decoding options can be used with regular beam search, prefix-constrained decoding, and/or random sampling. If the decoding option $return\_alternatives$ is used along with $target\_prefix$, the provided target left context is fed into the decoder in the teacher forcing mode,[6] then the engine expands the next $N$ most likely words, and continues (auto-completes) the decoding for these $N$ hypotheses independently. The shared task investigates four context cases:

---

[2] https://github.com/Helsinki-NLP/Tatoeba-Challenge
[3] https://github.com/google/sentencepiece
[4] https://github.com/fxsjy/jieba
[5] https://github.com/OpenNMT/CTranslate2/blob/master/docs/decoding.md
[6] In *teacher forcing* (Williams and Zipser, 1989), ground truth previous tokens are fed into the decoder, instead of the predicted tokens $y_{i-1}$ as suggested by Bahdanau et al. (2015)

(a) empty context, (b) right context only, (c) left context only, and (d) both the right and left contexts are provided. Hence, for all cases we returned multiple alternative translations, while for (c) and (d) we also returned another set of alternative auto-completions using the left context as a target prefix. In this sense, it is worth noting that we make use only of the left context, when available, and we do not use the right context at all, which we might investigate further in the future. To enhance diversity of translations, especially for (a) and (b), we applied random sampling with the CTranslate2's decoding option $sampling\_topk$, with various sampling temperatures. Our experiments are further elaborated in Section 4 and Section 5.

**Pinyin** The official Romanization system for Standard Mandarin Chinese is called Pinyin. Since the task organizers used the pypinyin library[7] to prepare the test files, we did too. OPUS English-to-Chinese models accept Chinese input, so we had to use the library to convert between the two writing systems. Since the conversion from Chinese characters to Pinyin is a lossy process and cannot be perfectly converted back, we keep a list of Chinese words resulted from tokenization with Jieba to be able to map Pinyin tokens to Chinese tokens later.

## 4 Method

We experimented with both beam search alternatives and random sampling, and found that the latter achieves better results. This could be due to the fact that alternatives generated from each beam are usually very similar, and lower beam values tend to generate translations of lower quality. This section elaborates on the actual methods we used for our submissions, while more details about initial experiments that led us to these decisions are explained in Section 5.

Random sampling is a decoding mode that randomly samples tokens from the model output distribution. In our experiments, we restrict the sampling to the top-10 candidates at each time-step. To obtain diverse generations from the MT model, we rely on randomness in the decoding method, in particular through top-K sampling that samples the next word from the top-K most probable choices

[7]https://github.com/mozillazg/python-pinyin

(Fan et al., 2018; Holtzman et al., 2018; Radford et al., 2019), instead of aiming to decode text that maximizes likelihood.

For each translation, we use the CTranlsate2 option *return_alternatives* to return 10 sequences, with 10 top-K sampling. If the entry has a left context starting with a capital letter, we use the prefix to constrain the decoding. In CTranslate2, combining *target_prefix* with the *return_alternatives* flag returns alternative sequences just after the prefix. We compose a list of alternatives with and without the prefix, and try to find the word starting with the typed sequence.[8] If the word is not found, we repeat the same process for up to five runs. In each new run, random sampling can generate a new set of alternatives. Our experiments show that returning 20 sequences with 20 top-K sampling could lead to more correctly predicted words (cf. Table 2); however, we had to consider the trade-off between quality and efficiency.[9]

Furthermore, we investigate increasing the randomness of the generation by using a value for sampling temperature between 1.0 and 1.3. For each run, a random value is generated in this range. The default sampling temperature in CTranslate2 is 1, which achieved relatively better results, as demonstrated in Table 1.

| Language | Settings | Accuracy | Human |
|---|---|---|---|
| de-en | ST=1.0 | 0.614441141 | 0.885 |
| | ST=1.3 | 0.609237735 | 0.8875 |
| en-de | ST=1.0 | 0.589418807 | 0.6725 |
| | ST=1.3 | 0.584939177 | 0.655 |
| zh-en | ST=1.0 + detok | 0.504113456 | 0.8675 |
| | ST=1.3 + detok | 0.502598878 | 0.8675 |
| | ST=1.0 | 0.493476989 | 0.86 |
| | ST=1.3 | 0.490619944 | 0.87 |
| en-zh | ST=1.0 | 0.319424091 | 0.5775 |
| | ST=1.3 | 0.319350821 | 0.5725 |

Table 1: Evaluation results on the test datasets. Automatic evaluation uses the "Accuracy" metric. "Human" refers to human evaluation. Results obtained from sampling temperature (ST) 1.0 are slightly better than those with the value 1.3. When the source is Chinese, detokenization (detok) resulted in slightly better scores.

[8]In a prefix-free target sequence, if multiple words start with the typed sequence, we return the first word. In practice, users could be prompted to choose from potential options.

[9]Our scripts are available at: https://github.com/ymoslem/WLAC

## 5 Other Experiments

This section elaborates on some initial experiments we conducted to decide what approach to use. The final approach we actually used in our submissions is explained in Section 4.

We used 10,000 entries of a Chinese-to-English golden sample provided by the organizers to evaluate various experiments. For sentence translation, when there is no left context, we experimented with the following values:

- beam size 1, 5, and 10, without sampling
- beam size 1, with random sampling top-K 10, 20, and 50

Table 2 shows the results for these experiments, and demonstrates that random sampling achieves the best overall accuracy. Random sampling with beam size 1 reveals better results than mere beam size 1 and even beam sizes 5 and 10 without random sampling. Multiple runs of random sampling can result in more correctly predicted words.

| Beam Size | Sampling Top-K | Hypotheses | Accuracy | Runs |
|---|---|---|---|---|
| 1 | N/A | 10 | 0.6519 | 1 |
| 5 | N/A | 10 | 0.6588 | 1 |
| 10 | N/A | 10 | 0.6573 | 1 |
| 1 | 10 | 10 | 0.6918 | 1 |
| 1 | 20 | 10 | 0.6907 | 1 |
| **1** | **20** | **20** | **0.7108** | **1** |
| 1 | 50 | 10 | 0.6853 | 1 |
| 5 | N/A | 10 | 0.6588 | 5 |
| 1 | 10 | 10 | 0.7165 | 5 |
| **1** | **20** | **20** | **0.7310** | **5** |

Table 2: Results for the Chinese-to-English golden sample dataset (10,000 entries). Random sampling outperforms even higher beam sizes.

## 6 API

Our API project[10] offers an easy way to integrate translation, auto-suggestion, and auto-completion features into translation environments. We chose FastAPI[11] for its high performance that beats many other Python web frameworks[12] in addition to its easy integration with OpenAPI (Swagger) documentation.

---
[10] https://github.com/ymoslem/SnowballMT
[11] https://github.com/tiangolo/fastapi
[12] https://www.techempower.com/benchmarks/#section=data-r20&hw=ph&test=query&l=zijzen-sf

### 6.1 API Endpoints

The API consists of a number of endpoints, receiving requests and sending the relevant responses in the JSON format. Each of the MT features has its endpoint.

#### 6.1.1 Translation Endpoint

The API handles a POST request (e.g. received from a CAT environment), including:

- $sentences$: list of the source sentences to be translated.
- $source\_language$: in a format like "$fr$" for French, and the default is "auto" to run language auto-detection.
- $target\_language$: in a format like "$en$" for English.

The API response is a list of strings for the MT translations in a JSON format.

#### 6.1.2 Auto-Suggestions Endpoint

When the user clicks on one word of the MT translation, the CAT environment sends a request to the API including:

- $sentence$: sentence to be translated.
- $prefix$: words to start the translation with.
- $source\_language$: in a format like "$fr$" for French, and the default is "auto" to run language auto-detection.
- $target\_language$: in a format like "$en$" for English.

The API response is a list of the MT word suggestions/alternatives for the current word, and the translation auto-completions if the user selects a specific suggestion.

### 6.2 JSON Response Examples

This is an example of a response to the translation request referred to in Section 6.1.1.

```
{ 'id': 10550004
  'source_lang': "fr",
  'target_lang': "en",
  'translations': [
    'The COVID-19 crisis has deepened already
        existing inequalities.'
  ]
}
```

This is an example of a response to the auto-suggestions request referred to in Section 6.1.2.

```
{
  'id': 10550005,
  'source_lang': "fr",
  'target_lang': "en",
  'result': {
    'translations': [
      {
        'suggestion': 'crisis',
        'compelection': 'of COVID-19 has deepened
            already existing inequalities.'
      },
      {
        'suggestion': 'COVID-19',
        'compelection': 'crisis has deepened already
            existing inequalities.'
      },
      {
        'suggestion': 'impact',
        'compelection': 'of COVID-19 crisis has
            deepened already existing inequalities
            .'
      }
    ]
  }
}
```

## Acknowledgements

This work is supported by the Science Foundation Ireland Centre for Research Training in Digitally-Enhanced Reality (d-real) under Grant No. 18/CRT/6224, the ADAPT Centre for Digital Content Technology which is funded under the Science Foundation Ireland (SFI) Research Centres Programme (Grant No. 13/RC/2106) and is co-funded under the European Regional Development Fund, and Microsoft Research.